\documentclass[10pt,twocolumn,letterpaper]{article}

\usepackage{iccv}
\usepackage{times}
\usepackage{epsfig}
\usepackage{graphicx}
\usepackage{caption}
\usepackage{amsmath}
\usepackage{amssymb}
\usepackage{multirow}
\usepackage{multicol}
\usepackage{longtable}
\usepackage{tablefootnote}

\usepackage[pagebackref=true,breaklinks=true,letterpaper=true,colorlinks,bookmarks=false]{hyperref}

\iccvfinalcopy 

\ificcvfinal\fi

\begin{document}

\title{FCOS3D: Fully Convolutional One-Stage Monocular 3D Object Detection\vspace{-1.0ex}}

\author{Tai Wang~~~~~~~~~~Xinge Zhu~~~~~~~~~~Jiangmiao Pang~~~~~~~~~~Dahua Lin\\
{\small CUHK-SenseTime Joint Lab, the Chinese University of Hong Kong}\\
{\small \{wt019, zx018, dhlin\}@ie.cuhk.edu.hk, pangjiangmiao@gmail.com}
}

\twocolumn[{%
\renewcommand\twocolumn[1][]{#1}%
\maketitle
\ificcvfinal\thispagestyle{empty}\fi
\begin{center}
  \centering
  \vspace{-1.0ex}
  \includegraphics[width=1.0\textwidth]{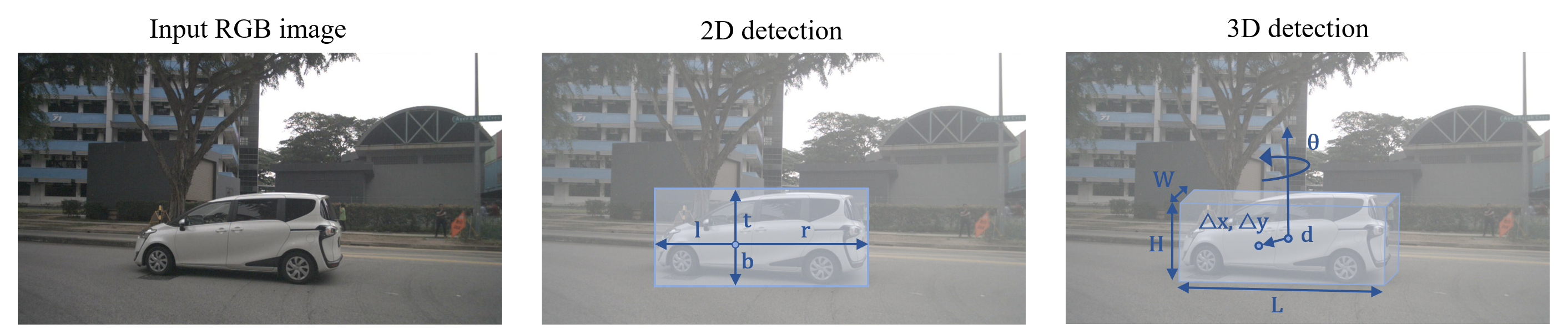}
  \vspace{-3.5ex}
  \captionof{figure}{Illustration of 2D detection and monocular 3D object detection. Given an input RGB image, a 2D anchor-free detector needs to predict the distance from a foreground point to four box sides. In contrast, a monocular 3D anchor-free detector needs to predict a transformed 3D-center, 3D size, and orientation of the object.
  }
  \label{fig:teaser}
\end{center}%
}]


\begin{abstract}
Monocular 3D object detection is an important task for autonomous driving considering its advantage of low cost. It is much more challenging than conventional 2D cases due to its inherent ill-posed property, which is mainly reflected in the lack of depth information.
Recent progress on 2D detection offers opportunities to better solving this problem. However, it is non-trivial to make a general adapted 2D detector work in this 3D task.
In this paper, we study this problem with a practice built on a fully convolutional single-stage detector and propose a general framework FCOS3D. Specifically, we first transform the commonly defined 7-DoF 3D targets to the image domain and decouple them as 2D and 3D attributes. Then the objects are distributed to different feature levels with consideration of their 2D scales and assigned only according to the projected 3D-center for the training procedure. Furthermore, the center-ness is redefined with a 2D Gaussian distribution based on the 3D-center to fit the 3D target formulation. All of these make this framework simple yet effective, getting rid of any 2D detection or 2D-3D correspondence priors.
Our solution achieves \textbf{1st place} out of all the vision-only methods in the nuScenes 3D detection challenge of NeurIPS 2020. Code and models are released at \footnotesize\url{https://github.com/open-mmlab/mmdetection3d}.
\end{abstract}
\vspace{-1.0ex}
\section{Introduction}
Object detection is a fundamental problem in computer vision. 
It aims to identify objects of interest in the image and predict their categories with corresponding 2D bounding boxes. 
With the rapid progress of deep learning, 2D object detection has been well explored in recent years.
Various models such as Faster R-CNN~\cite{FasterRCNN}, RetinaNet~\cite{RetinaNet}, and FCOS~\cite{FCOS} significantly promote the progress of the field and benefit various applications like autonomous driving.

However, 2D information is not enough for an intelligent agent to perceive the 3D real world. 
For example, when an autonomous vehicle needs to run smoothly and safely on the road, it must have accurate 3D information of objects around it to make secure decisions. Therefore, 3D object detection is becoming increasingly important in these robotic applications. 
Most state-of-the-art methods~\cite{VoxelNet,PointPillars,PointRCNN,reconfig_voxels,ssn,cylinder3d} rely on the accurate 3D information provided by LiDAR point clouds, but it is a heavy burden to install expensive LiDARs on each vehicle.
So monocular 3D object detection, as a simple and cheap setting for deployment, becomes a much meaningful research problem nowadays.

Considering monocular 2D and 3D object detection have the same input but different outputs, a straightforward solution for monocular 3D object detection is following the practices in the 2D domain but adding extra components to predict the additional 3D attributes of the objects.
Some previous work~\cite{MonoDIS,ROI10D} keeps predicting 2D boxes and further regresses 3D attributes on top of 2D centers and regions of interest. Others~\cite{M3D-RPN,D4LCN,Kinematic3D} simultaneously predict 2D and 3D boxes with 3D priors corresponding to each 2D anchor. Another stream of methods based on redundant 3D information~\cite{SS3D,RTM3D} predicts extra keypoints for optimized results ultimately.
In a word, the fundamental underlying problem is how to assign 3D targets to the 2D domain with the 2D-3D correspondence and predict them afterward.

In this paper, we adopt a simple yet efficient method to enable a 2D detector to predict 3D localization.
We first project the commonly defined 7-DoF 3D locations onto the 2D image and get the projected center point, which we name as 3D-center compared to the previous 2D-center.
With this projection, the 3D-center contains 2.5D information, \emph{i.e.}, 2D location and its corresponding depth. The 2D location can be further reduced to the 2D offset from a certain point on the image, which serves as the only 2D attribute that can be normalized among different feature levels like in the 2D detection. In comparison, depth, 3D size, and orientation are regarded as 3D attributes after decoupling. In this way, we transform the 3D targets with a center-based paradigm and avoid any necessary 2D detection or 2D-3D correspondence priors.

As a practical implementation, we build our method on FCOS~\cite{FCOS}, a simple anchor-free fully convolutional single-stage detector.
We first distribute the objects to different feature levels with consideration of their 2D scales.
Then the regression targets of each training sample are assigned only according to the projected 3D centers.
In contrast to FCOS that denotes the center-ness with distances to boundaries, we represent the 3D center-ness with a 2D Gaussian distribution based on the 3D-center.

We evaluate our method on a popular large-scale dataset, nuScenes~\cite{nuScenes}, and achieved \emph{1st place} on the camera track of this benchmark without any prior information. Moreover, we only need 2x less computing resources to train a baseline model with performance comparable to the previous best open-source method, CenterNet~\cite{CenterNet}, in one day, also 3x faster than it. Both show that our framework is simple and efficient. Detailed ablation studies show the importance of each component.

\label{sec:introduction}



\section{Related Work}
\begin{figure*}
\begin{center}
\includegraphics[width=1.0\linewidth]{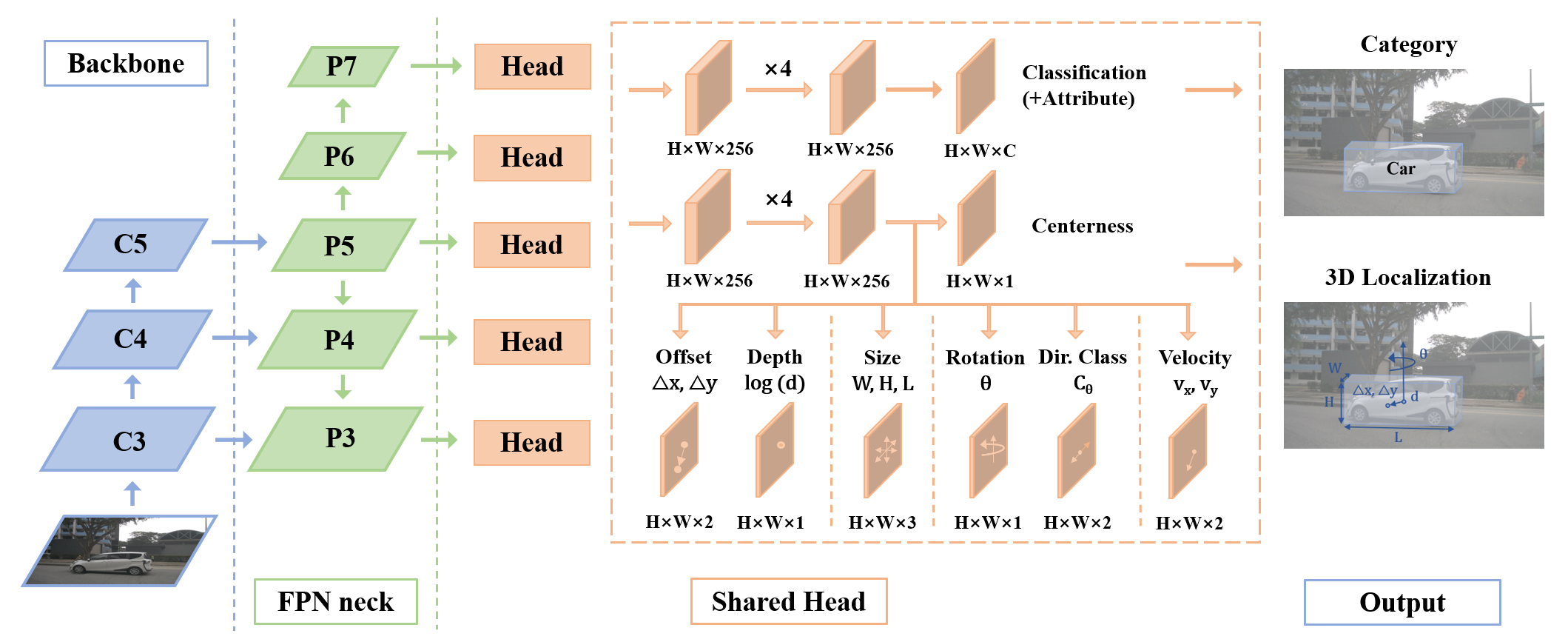}
\end{center}
   \vspace{-3ex}
   \caption{An overview of our pipeline. To leverage the well-developed 2D feature extractors, we basically follow the typical design of backbone and neck for 2D detectors. For detection head, we first reformulate the 3D targets with center-based paradigm to decouple it as multi-task learning. The strategies for multi-level target assignment and center sampling are further adjusted accordingly to equip this framework with the better capability of handling overlapped ground truths and scale variance problem.}
   \vspace{-2.5ex}
\label{fig: overview}
\end{figure*}
\noindent\textbf{2D Object Detection}\quad
Research on 2D object detection has made great progress with the breakthrough of deep learning approaches. According to the base of initial guesses, modern methods can be divided into two branches: anchor-based and anchor-free. Anchor-based methods~\cite{FastRCNN,FasterRCNN,SSD,YOLOv2} benefit from the predefined anchors in terms of much easier regression while having many hyper-parameters to tune. In contrast, anchor-free methods~\cite{DenseBox,YOLOv1,FCOS,CornerNet,CenterNet} do not need these prior settings and are thus neater with better universality. For simplicity, this paper takes FCOS, a representative anchor-free detector, as the baseline considering its capability of handling overlapped ground truths and scale variance problem.

From another perspective, monocular 3D detection is a more difficult task closely related to 2D detection. But there is few work investigating the connection and difference between them, which makes them isolated and not able to benefit from the advancement of each other. This paper aims to adapt FCOS as the example and further build a closer connection between these two tasks.

\noindent\textbf{Monocular 3D Object Detection}\quad
Monocular 3D detection is more complex than conventional 2D detection. The underlying key problem is the inconsistency of input 2D data modal and the output 3D predictions.

\noindent\emph{Methods involving sub-networks}\quad The first batch of works resorts to sub-networks to assist 3D detection. To mention only a few, 3DOP~\cite{3DOP} and MLFusion~\cite{MLFusion} use a depth estimation network, while Deep3DBox~\cite{Deep3DBox} uses a 2D object detector. They heavily rely on the performance of sub-networks, even external data and pre-trained models, making the entire system complex and inconvenient to train.

\noindent\emph{Transform to 3D representations}\quad Another category of methods converts the input RGB image to other 3D representations, such as voxels~\cite{OFTNet} and point clouds~\cite{PseudoLiDAR}. Recent work~\cite{PseudoLiDAR++,End2EndPL,Foresee,CaDDN} has made great progress following this approach and shown promising performance. However, they still rely on dense depth labels and thus are not regarded as pure monocular approaches. There are also domain gaps between different depth sensors and LiDARs, making them hard to generalize to new practice settings smoothly. In addition, it is difficult to process a large number of point clouds when applying these methods to the real-world scenarios.

\noindent\emph{End-to-end design like 2D detection}\quad Recent work notices these drawbacks and begins to design end-to-end frameworks like 2D detectors. For example, M3D-RPN~\cite{M3D-RPN} proposes a single-stage detector with an end-to-end region proposal network and depth-aware convolution. SS3D~\cite{SS3D} detects 2D key points and further predicts object characteristics with uncertainties. MonoDIS~\cite{MonoDIS} improves the multi-task learning with a disentangling loss. These methods follow the anchor-based manners and are thus required to define consistent 2D and 3D anchors. Some of them also need multiple training stages or hand-crafted post-optimization phases. In contrast, anchor-free methods~\cite{CenterNet,RTM3D,MonoPair} do not need to make statistics on the given data. It is more convenient to generalize their simple designs to more complex cases with more various classes or different intrinsic settings. Hence, we choose to follow this paradigm.


Nevertheless, these works hardly study the key difficulty when applying a general 2D detector to monocular 3D detection. What should be kept or adjusted therein is seldom discussed when proposing their new frameworks. In contrast, this paper concentrates on this point, which could provide a reference when applying a typical 2D detector framework to a closely related task. On this basis, a more in-depth understanding of the connection and difference between these two tasks will also benefit further research of both communities.

\label{sec:related}


\section{Approach}
\label{sec:approach}
Object detection is one of the most fundamental and challenging problems for scene understanding. The goal of conventional 2D object detection is to predict 2D bounding boxes and category labels for each object of interest. In comparison, monocular 3D detection needs us to predict 3D bounding boxes instead, which need to be decoupled and transformed to the 2D image plane. This section will first present an overview of our framework with our adopted reformulation of 3D targets, and then elaborate on two corresponding technical designs, 2D guided multi-level 3D prediction and 3D center-ness with 2D Gaussian distribution, tailored to this task. These technical designs work together to equip the 2D detector FCOS with the capability of detecting 3D objects.

\subsection{Framework Overview}
A fully convolutional one-stage detector typically consists of three components: a backbone for feature extraction, necks for multi-level branches construction and detection heads for dense predictions. Then we briefly introduce each of them.

\noindent\textbf{Backbone}\quad We use the pretrained ResNet101~\cite{ResNet,ImageNet} with deformable convolutions~\cite{DeformConv} for feature extraction. It achieves a good trade-off between accuracy and efficiency in our experiments. We fixed the parameters of the first convolutional block to avoid more memory overhead.

\noindent\textbf{Neck}\quad The second module is the Feature Pyramid Network~\cite{FPN}, a primary component for detecting objects at different scales. For precise clarification, we denote feature maps from level 3 to 7 as P3 to P7, as shown in Fig.~\ref{fig: overview}. We follow the original FCOS to obtain P3 to P5 and downsample P5 with two convolutional blocks to obtain P6 and P7. All of these five feature maps are responsible for predictions of different scales afterward.

\noindent\textbf{Detection Head}\quad Finally, for shared detection heads, we need to deal with two critical issues. The first is how to distribute targets to different feature levels and different points. It is one of the core problems for different detectors and will be presented in Sec.~\ref{sec: target_assign}. The second is how to design the architecture. We follow the conventional design of RetinaNet~\cite{RetinaNet} and FCOS~\cite{FCOS}. Each shared head consists of 4 shared convolutional blocks and small heads for different targets. It is empirically more effective to build extra disentangled heads for \emph{regression} targets with different measurements, so we set one small head for each of them (Fig.~\ref{fig: overview}).

So far, we have introduced the overall design of our network architecture. Next, we will formulate this problem more formally and present the detailed training and inference procedure.

\noindent\textbf{Regression Targets}\quad To begin with, we first recall the formulation of anchor-free manners for object detection in FCOS. Given a feature map at layer $i$ of the backbone, denoted as $F_i\in\mathbb{R}^{H\times W\times C}$, we need to predict objects based on each point on this feature map, which corresponds to uniformly distributed points on the original input image. Formally, for each location $(x, y)$ on the feature map $F_i$, suppose the total stride until layer $i$ is $s$, then the corresponding location on the original image should be $(sx+\lfloor\frac{s}{2}\rfloor, sy+\lfloor\frac{s}{2}\rfloor)$. Unlike anchor-based detectors regressing targets by taking predefined anchors as a reference, we directly predict objects based on these locations. Moreover, because we do not rely on anchors, the criterion for judging whether a point is from the foreground or not will no longer be the IoU (Intersection over Union) between anchors and ground truths. Instead, as long as the point is near the box center enough, it could be a foreground point.

\begin{figure}
\begin{center}
\includegraphics[width=1.0\linewidth]{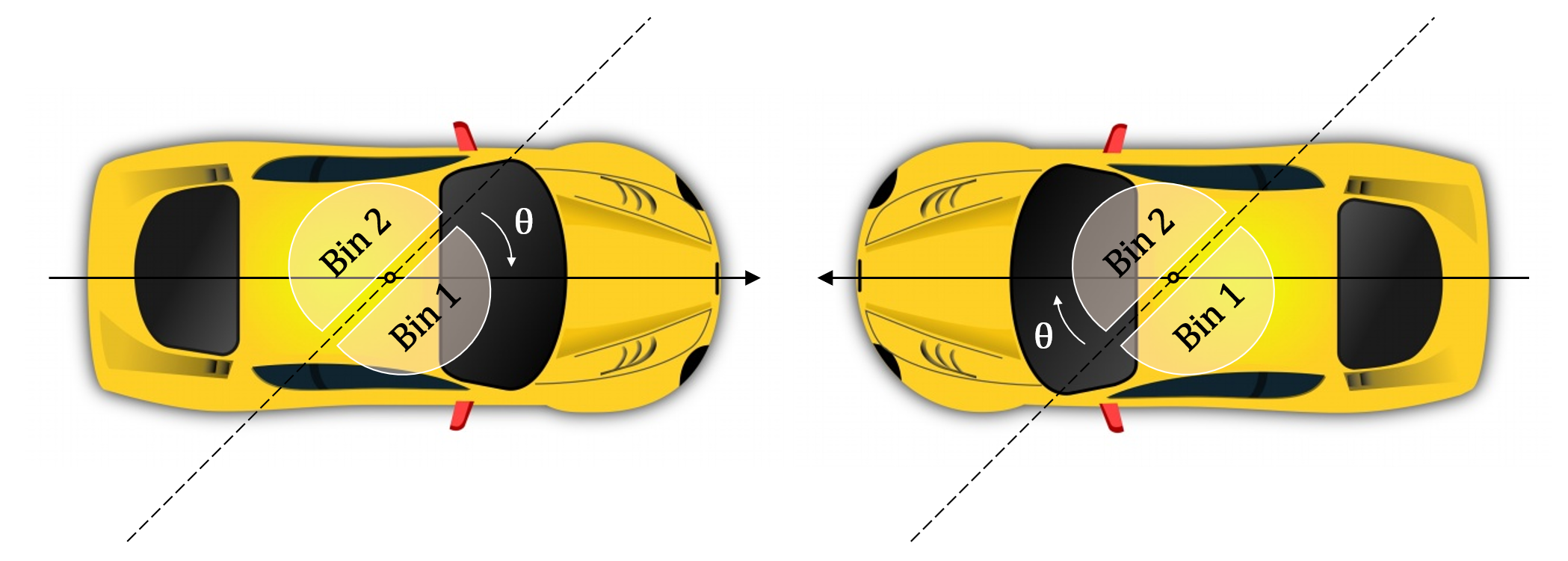}
\end{center}
   \vspace{-3.5ex}
   \caption{Our exploited rotation encoding scheme. Two objects with opposite orientations share the same rotation offset based on the 2-bin boundary, thus have the same $sin$ value. To distinguish them, we predict an additional direction class from the regression branch.}
   \vspace{-4ex}
\label{fig: rot_encoding}
\end{figure}
In the 2D case, the model needs to regress the distance of the point to the top/bottom/left/right side, denoted as $t, b, l, r$ in Fig.~\ref{fig:teaser}. However, in the 3D case, it is non-trivial to regress the distance to six faces of the 3D bounding box. Instead, a more straightforward implementation is to convert the commonly defined 7-DoF regression targets to the 2.5D center and 3D size. The 2.5D center can be easily transformed back to 3D space with a camera intrinsic matrix. Regressing the 2.5D center could be further reduced to regressing the offset from the center to a specific foreground point, $\Delta x, \Delta y$, and its corresponding depth $d$ respectively. In addition, to predict the allocentric orientation of the object, we divide it into two parts: angle $\theta$ with period $\pi$ and 2-bin direction classification. The first component naturally models the IOU of our predictions with the ground truth boxes, while the second component focuses on the adversarial case where two boxes have opposite orientations. Benefiting from this angle encoding, our method surpasses another center-based framework, CenterNet, in terms of orientation accuracy, which will be compared in the experiments. The rotation encoding scheme is illustrated in Fig.~\ref{fig: rot_encoding}.

In addition to these regression targets related to the location and orientation of objects, we also regress a binary target center-ness $c$ like FCOS. It serves as a soft binary classifier to determine which points are closer to centers, and helps suppress those low-quality predictions far away from object centers. More details are presented in Sec.~\ref{sec:centerness}.

To sum up, the regression branch needs to predict $\Delta x, \Delta y, d, w, l, h, \theta, v_x, v_y$, direction class $C_\theta$ and center-ness $c$ while the classification branch needs to output the class label of the object and its attribute label (Fig.~\ref{fig: overview}).

\noindent\textbf{Loss}\quad For classification and different regression targets, we define their loss respectively and take their weighted summation as the total loss. Firstly, for classification branch, we use the commonly used focal loss~\cite{RetinaNet} for object classification loss:
\vspace{-1ex}
\begin{equation}
    \centering
    L_{cls} = -\alpha(1-p)^\gamma logp
    \vspace{-1ex}
\end{equation}
where $p$ is the class probability of a predicted box. We follow the settings, $\alpha = 0.25$ and $\gamma = 2$, of the original paper. For attribute classification, we use a simple softmax classification loss, denoted as $L_{attr}$.\\
For regression branch, we use smooth L1 loss for each regression targets except center-ness with corresponding weights considering their scales:
\vspace{-1ex}
\begin{equation}
    \label{eqn: loc_loss}
    \centering
    L_{loc} = \sum_{b\in (\Delta x, \Delta y, d, w, l, h, \theta, v_x, v_y)} SmoothL1(\Delta b)
    \vspace{-1ex}
\end{equation}
where the weight of $\Delta x, \Delta y, w, l, h, \theta$ error is 1, the weight of $d$ is 0.2 and the weight of $v_x, v_y$ is 0.05. Note that although we employ $exp(x)$ for depth prediction, we still compute the loss in the original depth space instead of the log space. It empirically results in more accurate depth estimation ultimately. We use the softmax classification loss and binary cross entropy (BCE) loss for direction classification and center-ness regression, denoted as $L_{dir}$ and $L_{ct}$ respectively. Finally, the total loss is:
\begin{equation}
    \centering\footnotesize
    L = \frac{1}{N_{pos}}(\beta_{cls}L_{cls}+\beta_{attr}L_{attr}+\beta_{loc}L_{loc}+\beta_{dir}L_{dir}+\beta_{ct}L_{ct})
\end{equation}
where $N_{pos}$ is the number of positive predictions and $\beta_{cls} = \beta_{attr} = \beta_{loc} = \beta_{dir} = \beta_{ct} = 1$.

\noindent\textbf{Inference}\quad During inference, given an input image, we forward it through the framework and obtain bounding boxes with their class scores, attribute scores, and center-ness predictions. We multiply the class score and center-ness as the confidence for each prediction and conduct rotated Non-Maximum Suppression (NMS) in the bird view as most 3D detectors to get the final results.

\begin{figure}
\begin{center}
\includegraphics[width=1.0\linewidth]{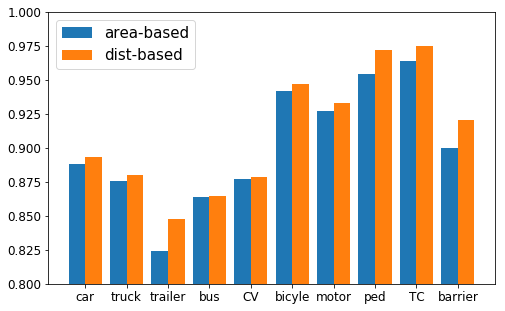}
\end{center}
   \vspace{-3ex}
   \caption{Our proposed distance-based target assignment for dealing with ambiguity case could significantly improve the best possible recall (BPR) for each class, especially for large objects like trailers. Construction vehicle and traffic cone are abbreviated as CV and TC in this figure.}
   \vspace{-3ex}
\label{fig: target_assign}
\end{figure}

\subsection{2D Guided Multi-Level 3D Prediction}\label{sec: target_assign}
As mentioned previously, to train a detector with pyramid networks, we need to devise a strategy to distribute targets to different feature levels. FCOS~\cite{FCOS} has discussed two crucial issues therein: 1) How to enable anchor-free detectors to achieve similar Best Possible Recall (BPR) compared to anchor-based methods, 2) Intractable ambiguity problem caused by overlaps of ground-truth boxes. The comparison in the original paper has well addressed the first problem. It shows that multi-level prediction through FPN can improve BPR and even achieve better results than anchor-based methods. Similarly, the conclusion of this problem is also applicable in our adapted framework. The second question will involve the specific setting of the regression target, which we will discuss next.

The original FCOS detects objects of different sizes in different levels of feature maps. Different from anchor-based methods, instead of assigning anchors with different sizes, it directly assigns ground-truth boxes with different sizes to different levels of feature maps. Formally, it first computes the 2D regression targets, $l^*$, $r^*$, $t^*$, $b^*$  for each location at each feature level. Then locations satisfying $max(l^*, r^*, t^*, b^*)>m_i$ or $max(l^*, r^*, t^*, b^*)<m_{i-1}$ would be regarded as a negative sample, where $m_i$ denotes the maximum regression range for feature level $i$~\footnote{We set the regression range as (0, 48, 96, 192, 384, $\infty$) for $m_2$ to $m_7$ in our experiments respectively.}. In comparison, we also follow this criterion in our implementation, considering that the scale of 2D detection is directly consistent with how large a region we need to focus on. However, we only use 2D detection for filtering meaningless targets in this assignment step. After completing the target assignment, our regression targets only include 3D-related ones. Here we generate the 2D bounding boxes by computing the exterior rectangle of projected 3D bounding boxes, so we do not need any 2D detection annotations or priors.

Next, we will discuss how to deal with the ambiguity problem. Specifically, when a point is inside multiple ground truth boxes in the same feature level, which box should be assigned to it? The usual way is to select according to the area of the 2D bounding box. The box with a smaller area is selected as the target box for this point. We call this scheme the \emph{area-based} criterion. This scheme has an obvious drawback: Large objects will be paid less attention by such processing, which is also verified by our experiments (Fig.~\ref{fig: target_assign}). Taking this into account, we instead propose a \emph{distance-based} criterion, \emph{i.e.}, select the box with closer center as the regression target. This scheme is consistent with the adapted center-based mechanism for defining regression targets. Furthermore, it is also reasonable because the points closer to the object's center can obtain more comprehensive and balanced local region features, thus easily producing higher-quality predictions.
Through simple verification (Fig.~\ref{fig: target_assign}), we find that this scheme significantly improves the best possible recall (BPR) and mAP of large objects and also improves the overall mAP (about 1\%), which will be presented in the ablation study.

In addition to the center-based approach to deal with ambiguity, we also use the 3D-center to determine foreground points, \emph{i.e.}, only the points near the center enough will be regarded as positive samples. We define a hyper-parameter, radius, to measure this central portion. The points with a distance smaller than radius$\times$stride to the object center would be considered positive, where the radius is set to 1.5 in our experiments.

Finally, we replace each output $x$ of different regression branches with $s_ix$ to distinguish shared heads for different feature levels. Here $s_i$ is a trainable scalar used to adjust the exponential function base for feature level $i$. It brings a minor improvement in terms of detection performance.

\begin{table*}
\small
    \caption{Results on the nuScenes dataset.}
    \vspace{-4ex}
	\begin{center}
	\begin{tabular}{c|c|c|c|c|c|c|c|c|c}
	\hline
	Methods & Dataset & Modality & mAP & mATE & mASE & mAOE & mAVE & mAAE & NDS\\
	\hline\hline
	CenterFusion~\cite{CenterFusion} & test & Camera \& Radar & 0.326 & 0.631 & 0.261 & 0.516 & 0.614 & 0.115 & 0.449\\
	\hline
	PointPillars~\cite{PointPillars} & test & LiDAR & 0.305 & 0.517 & 0.290 & 0.500 & 0.316 & 0.368 & 0.453\\
	\hline
	MEGVII~\cite{CBGS} & test & LiDAR & \textbf{0.528} & 0.300 & 0.247 & 0.379 & 0.245 & 0.140 & \textbf{0.633}\\
	\hline\hline
	LRM0 & test & Camera & 0.294 & 0.752 & 0.265 & 0.603 & 1.582 & 0.14 & 0.371\\
	\hline
	MonoDIS~\cite{MonoDIS} & test & Camera & 0.304 & 0.738 & 0.263 & 0.546 & 1.553 & 0.134 & 0.384\\
	\hline
	CenterNet~\cite{CenterNet} (HGLS) & test & Camera & 0.338 & 0.658 & 0.255 & 0.629 & 1.629 & 0.142 & 0.4\\
	\hline
	Noah CV Lab & test & Camera & 0.331 & 0.660 & 0.262 & 0.354 & 1.663 & 0.198 & 0.418\\
	\hline
	FCOS3D (Ours) & test & Camera & \textbf{0.358} & 0.690 & 0.249 & 0.452 & 1.434 & 0.124 & \textbf{0.428}\\
	\hline\hline
	CenterNet~\cite{CenterNet} (DLA) & val & Camera & 0.306 & 0.716 & 0.264 & 0.609 & 1.426 & 0.658 & 0.328\\
	\hline
	FCOS3D (Ours) & val & Camera & \textbf{0.343} & 0.725 & 0.263 & 0.422 & 1.292 & 0.153 & \textbf{0.415}\\
	\hline
	\end{tabular}
	\end{center}
	\vspace{-5.0ex}
	\label{tab: quantitative}
\end{table*}

\subsection{3D Center-ness with 2D Gaussian Distribution}\label{sec:centerness}
In the original design of FCOS, center-ness $c$ is defined by 2D regression targets, l*, r*, t*, b*:
\vspace{-1ex}
\begin{equation}
    \centering
    c = \sqrt{\frac{min(l^*, r^*)}{max(l^*, r^*)}\times \frac{min(t^*, b^*)}{max(t^*, b^*)}}
    \vspace{-1ex}
\end{equation}
Because our regression targets are changed to the 3D center-based paradigm, we define the center-ness by 2D Gaussian distribution with the projected 3D-center as the origin. The 2D Gaussian distribution is simplified as:
\vspace{-1ex}
\begin{equation}
    \centering
    c = e^{-\alpha ((\Delta x)^2+(\Delta y)^2)}
    \vspace{-1ex}
\end{equation}
Here $\alpha$ is used to adjust the intensity attenuation from the center to the periphery and set to 2.5 in our experiments. We take it as the ground truth of center-ness and predict it from the regression branch for filtering low-quality predictions later. As mentioned earlier, this center-ness target ranges from 0 to 1, so we use the Binary Cross Entropy (BCE) loss for training that branch.

\section{Experimental Setup}
\label{sec:experimental_setup}
\subsection{Dataset}
We evaluate our framework on a large-scale, commonly used dataset, nuScenes~\cite{nuScenes}. It consists of multi-modal data collected from 1000 scenes, including RGB images from 6 surround-view cameras, points from 5 Radars and 1 LiDAR. It is split into 700/150/150 scenes for training/validation/testing. There are overall 1.4M annotated 3D bounding boxes from 10 categories. Due to its variety of scenes and ground truths, it is becoming one of the authoritative benchmarks for 3D object detection. Therefore, we take it as the platform to validate the efficacy of our method.

\subsection{Evaluation Metrics}
We use the official metrics, distance-based mAP, and NDS for a fair comparison with other methods. Next, we briefly introduce these two kinds of metrics as follows.

\noindent\textbf{Average Precision metric} \quad
The Average Precision (AP) metric is generally used when evaluating the performance of object detectors. Instead of using 3D Intersection over Union (IoU) for thresholding, nuScenes defines the match by 2D center distance $d$ on the ground plane for decoupling detection from object size and orientation. On this basis, we calculate AP by computing the normalized area under the precision-recall curve for recall and precision over 10\%. Finally, mAP is computed over all matching thresholds, $\mathbb{D} = \{0.5, 1, 2, 4\}$ meters, and all categories $\mathbb{C}$:
\vspace{-1ex}
\begin{equation}
    \centering
    mAP = \frac{1}{|\mathbb{C}||\mathbb{D}|}\sum_{c\in\mathbb{C}}\sum_{d\in\mathbb{D}}AP_{c,d}
    \vspace{-0.6ex}
\end{equation}
\noindent\textbf{True Positive metrics} \quad
Apart from Average Precision, we also calculate five kinds of True Positive metrics, Average Translation Error (ATE), Average Scale Error (ASE), Average Orientation Error (AOE), Average Velocity Error (AVE) and Average Attribute Error (AAE). To obtain these measurements, we firstly define that predictions with center distance from the matching ground truth $d \le 2m$ will be regarded as true positives (TP). Then matching and scoring are conducted independently for each class of objects, and each metric is the average cumulative mean at each recall level above 10\%. ATE is the Euclidean center distance in 2D ($m$). ASE is equal to $1-IOU$, $IOU$ is calculated between predictions and labels after aligning their translation and orientation. AOE is the smallest yaw angle difference between predictions and labels ($radians$). Note that different from other classes measured on the entire $360^{\circ}$ period, barriers are measured on $180^{\circ}$ period. AVE is the L2-Norm of the absolute velocity error in 2D ($m/s$). AAE is defined as $1-acc$, where $acc$ refers to the attribute classification accuracy. Finally, given these metrics, we compute the mean TP metric (mTP) overall all categories:
\vspace{-1ex}
\begin{equation}
    \centering
    mTP = \frac{1}{|\mathbb{C}|}\sum_{c\in\mathbb{C}}TP_c
    \vspace{-1ex}
\end{equation}
Note that not well-defined metrics will be omitted, like AVE for cones and barriers, considering they are stationary.

\noindent\textbf{NuScenes Detection Score} \quad
The conventional mAP couples the evaluation of locations, sizes, and orientations of detections and also could not capture some aspects in this setting like velocity and attributes, so this benchmark proposes a more comprehensive, decoupled but simple metric, nuScenes detection score (NDS):
\vspace{-1ex}
\begin{equation}
    \centering
    NDS = \frac{1}{10}[5mAP+\sum_{mTP\in\mathbb{TP}}(1-min(1, mTP))]
    \vspace{-1ex}
\end{equation}
where mAP is mean Average Precision (mAP) and $\mathbb{TP}$ is the set composed of five True Positive metrics. Considering mAVE, mAOE and mATE can be larger than 1, a bound is applied to limit them between 0 and 1.

\begin{table*}
\small
    \caption{Average precision for each class on the nuScenes test benchmark. CV and TC are abbreviation of construction vehicle and traffic cone in the table.}
    \vspace{-3.5ex}
	\begin{center}
	\begin{tabular}{c|c|c|c|c|c|c|c|c|c|c|c}
	\hline
	Methods & car & truck & bus & trailer & CV & ped & motor & bicycle & TC & barrier & mAP\\
	\hline\hline
	LRM0 & 0.467 & 0.21 & 0.17 & 0.149 & 0.061 & 0.359 & 0.287 & 0.246 & 0.476 & 0.512 & 0.294\\
	\hline
	MonoDIS~\cite{MonoDIS} & 0.478 & 0.22 & 0.188 & 0.176 & 0.074 & 0.37 & 0.29 & 0.245 & 0.487 & 0.511 & 0.304\\
	\hline
	CenterNet~\cite{CenterNet} (HGLS) & 0.536 & 0.27 & 0.248 & 0.251 & 0.086 & 0.375 & 0.291 & 0.207 & 0.583 & 0.533 & 0.338\\
	\hline
	Noah CV Lab & 0.515 & 0.278 & 0.249 & 0.213 & 0.066 & 0.404 & 0.338 & 0.237 & 0.522 & 0.49 & 0.331\\
	\hline
	FCOS3D (Ours) & 0.524 & 0.27 & 0.277 & 0.255 & 0.117 & 0.397 & 0.345 & 0.298 & 0.557 & 0.538 & \textbf{0.358}\\
	\hline
	\end{tabular}
	\end{center}
	\vspace{-3.0ex}
	\label{tab: ap_class}
\end{table*}

\begin{table*}
\small
    \caption{Ablation studies on the nuScenes validation 3D detection benchmark.}
    \vspace{-3.5ex}
	\begin{center}
	\begin{tabular}{c|c|c|c|c|c|c|c}
	\hline
	Methods & mAP & mATE & mASE & mAOE & mAVE & mAAE & NDS\\
	\hline\hline
	Baseline (FCOS + 3D targets) & 0.227 & 0.868 & 0.272 & 0.778 & 1.326 & 0.393 & 0.282\\
	\hline
	+ Depth loss in original space & 0.25 & 0.838 & 0.268 & 0.892 & 1.33 & 0.413 & 0.284\\
	\hline
	+ Flip augmentation & 0.248 & 0.85 & 0.267 & 1.016 & 1.358 & 0.268 & 0.286\\
	\hline
	+ Dist-based target assign  \& attr pred & 0.257 & 0.832 & 0.268 & 0.852 & 1.2 & 0.18 & 0.316\\
	\hline
	+ NMS among predictions of six views & 0.26 & 0.828 & 0.267 & 0.85 & 1.371 & 0.18 & 0.317\\
	\hline
	+ Stronger backbone (ResNet101) & 0.272 & 0.821 & 0.265 & 0.81 & 1.379 & 0.17 & 0.329\\
	\hline
	+ Disentangled heads & 0.28 & 0.822 & 0.274 & 0.64 & 1.305 & 0.177 & 0.349\\
	\hline
	+ DCN in backbone & 0.295 & 0.806 & 0.268 & 0.511 & 1.315 & 0.17 & 0.372\\
	\hline
	+ Finetune w/ depth weight=1.0 & 0.316 & 0.755 & 0.263 & 0.458 & 1.307 & 0.169 & 0.393\\
	\hline
	+ Test time augmentation & 0.326 & 0.743 & 0.259 & 0.441 & 1.341 & 0.163 & 0.402\\
	\hline
	+ More epochs \& ensemble & \textbf{0.343} & 0.725 & 0.263 & 0.422 & 1.292 & 0.153 & \textbf{0.415}\\
	\hline
	\end{tabular}
	\end{center}
	\vspace{-4.5ex}
	\label{tab: ablation}
\end{table*}

\subsection{Implementation Details}
\noindent\textbf{Network Architectures}\quad
As shown in Fig.~\ref{fig: overview}, our framework follows the design of FCOS. Given the input image, we utilize ResNet101 as the feature extraction backbone followed by Feature Pyramid Networks (FPN) for generating multi-level predictions. Detection heads are shared among multi-level feature maps except that three scale factors are used to differentiate some of their final regressed results, including offsets, depths, and sizes, respectively. All the convolutional modules are made up of basic convolution, batch normalization, and activation layers, and normal distribution is leveraged for weights initialization. The overall framework is built on top of MMDetection3D~\cite{mmdet3d}.

\noindent\textbf{Training Parameters}\quad
For all experiments, we trained randomly initialized networks from scratch following end-to-end manners. Models are trained with an SGD optimizer. Gradient clip and warm-up policy are exploited with the learning rate 0.002, the number of warm-up iterations 500, warm-up ratio 0.33, and batch size 32 on 16 GTX 1080Ti GPUs. We apply a weight of 0.2 for depth regression to train our baseline model to make the training more stable. For a more competitive performance and a more accurate detector, we finetune our model with this weight switched to 1. Related results are presented in the ablation study.

\noindent\textbf{Data Augmentation}\quad
Like previous work, we only implement image flip for data augmentation both when training and testing. Note that only the offset is needed to be flipped as 2D attributes and 3D boxes need to be transformed correspondingly in 3D space when flipping images. For test time augmentation, we average the score maps output by the detection heads except rotation and velocity related scores due to their inaccuracy. It is empirically a more efficient approach for augmentation than merging boxes at last.

\begin{figure*}
\begin{center}
\includegraphics[width=1.0\linewidth]{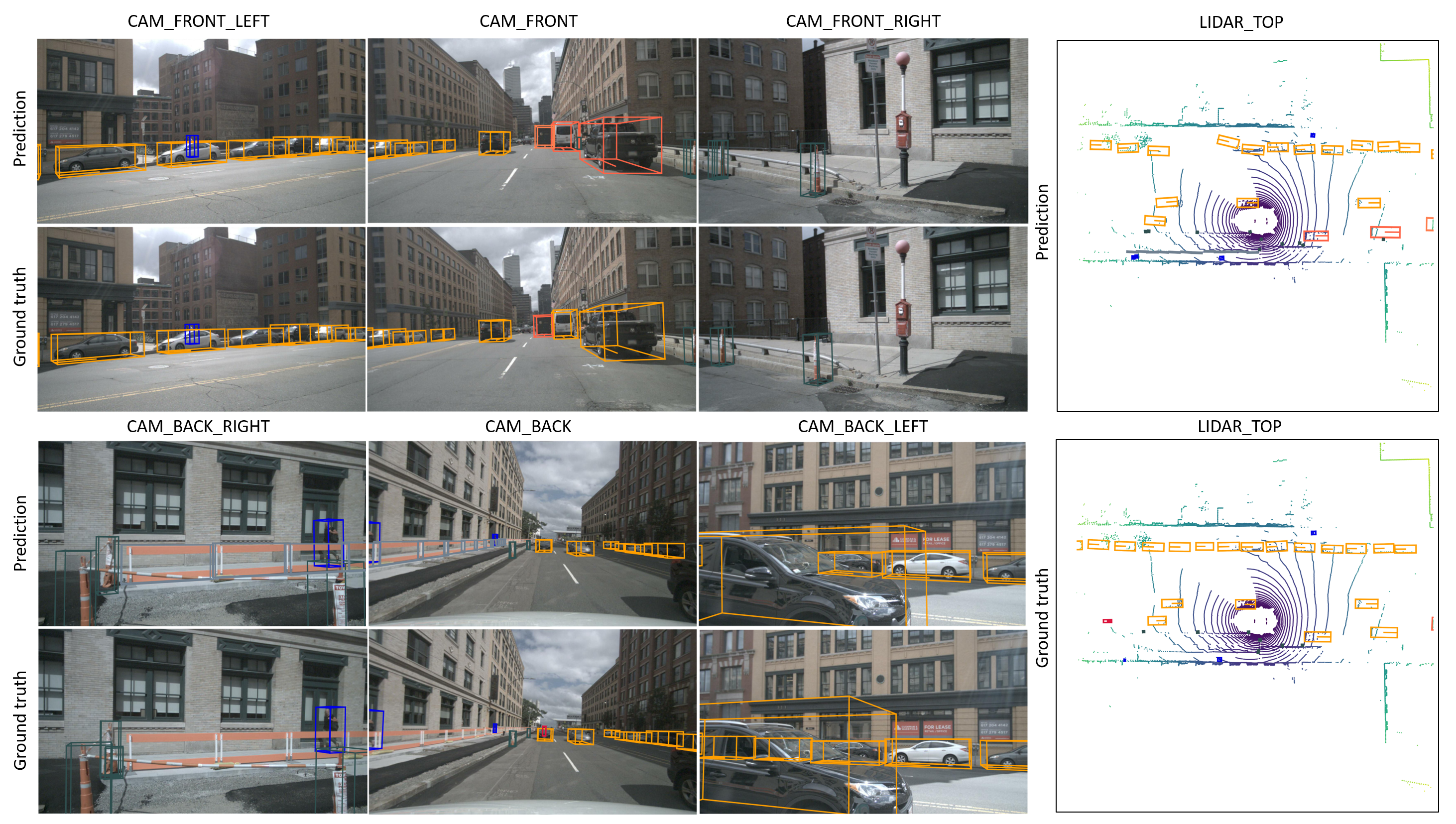}
\end{center}
   \vspace{-3.2ex}
   \caption{Qualitative analysis of detection results. 3D bounding boxes predictions are projected onto images from six different views and bird-view, respectively. Boxes from different categories are marked with different colors. We can see that the results are reasonable except for some detection with false class predictions from the left part. Moreover, a few small objects are detected by our model while not annotated as ground truth, like barriers in the back/back right camera. However, apart from the inherent occlusion problem in this setting, depth and orientation estimations of some objects are still inaccurate, which can be observed in the visualization from bird-view.}
   \vspace{-2.8ex}
\label{fig: qualitative}
\end{figure*}

\section{Results}
In this section, we present quantitative and qualitative results and make a detailed ablation study on essential factors in pushing our method towards the state-of-the-art.
\label{sec:results}
\subsection{Quantitative Analysis}
First, we show the results of quantitative analysis in Tab.~\ref{tab: quantitative}. We compare the results on the test set and validation set, respectively. We first compared all the methods using RGB images as the input data on the test set. We achieved the best performance among them with mAP 0.358 and NDS 0.428. In particular, our method exceeded the previous best one by more than 2\% in terms of mAP. Benchmarks using LiDAR data as the input include PointPillars~\cite{PointPillars}, which are faster and lighter, and CBGS~\cite{CBGS} (MEGVII in the Tab.~\ref{tab: quantitative}) with relatively high performance. For the approaches which use the input of RGB image and Radar data, we select CenterFusion~\cite{CenterFusion} as the benchmark. It can be seen that although our method has a certain gap with the high-performance CBGS, it even surpasses PointPillars and CenterFusion on mAP. It shows that we can solve this ill-posed problem decently with enough data. At the same time, it can be seen that the methods using other modals of data have relatively better NDS, mainly because the mAVE is smaller. The reason is that other methods introduce continuous multi-frame data, such as point cloud data from consecutive frames, to predict the speed of objects. In addition, Radars can measure the velocity, so CenterFusion can achieve reasonable speed prediction even with a single frame image. However, these can not be achieved with only a single image, so how to mine the speed information from consecutive frame images will be one of the directions that can be explored in the future. For detailed mAP for each category, please refer to Tab.~\ref{tab: ap_class} and the official benchmark.

On the validation set, we compare our method with the best open-source detector, CenterNet. Their method not only takes about three days to train (compared with our only one day to achieve comparable performance, possibly thanks to our pre-trained backbone) but also is inferior to our method except for mATE. In particular, thanks to our rotation encoding scheme, we achieved a significant improvement in the accuracy of angle prediction. The significant improvement of mAP reflects the superiority of our multi-level prediction. Based on all the improvements in these aspects, we finally achieved a gain of about 9\% on NDS.

\subsection{Qualitative Analysis}
Then we show some qualitative results in Fig.~\ref{fig: qualitative} to give an intuitive understanding of the performance of our model. First of all, in Fig.~\ref{fig: qualitative}, we draw the predicted 3D bounding boxes in the six-view images and the top-view point clouds. For example, the barriers in the camera at the rear right are not labeled but detected by our model. However, at the same time, we should also see that our method still has apparent problems in the depth estimation and identification of occluded objects. For example, it is difficult to detect the blocked car in the left rear image. Moreover, from the top view, especially in terms of depth estimation, results are not as good as those shown in the image. This is also in line with our expectation that depth estimation is still the core challenge in this ill-posed problem.

\subsection{Ablation Studies}
Finally, we show some critical factors in the whole process of studying in Tab.~\ref{tab: ablation}. It can be seen that in the prophase process, transforming depth back to the original space to compute loss is an essential factor to improve mAP, and distance-based target assignment is an essential factor to improve the overall NDS. The stronger backbone, such as replacing the original ResNet50 with ResNet101 and using DCN, is crucial in the later promotion process. At the same time, due to the difference in scales and measurements, using disentangled heads for different regression targets is also a meaningful way to improve the accuracy of angle prediction and NDS. Finally, we achieve the current state-of-the-art through simple augmentation, more training epochs, and a basic model ensemble.
\section{Conclusion}
\label{sec:conclusion}
This paper proposes a simple yet efficient one-stage framework, FCOS3D, for monocular 3D object detection without any 2D detection or 2D-3D correspondence priors. In the framework, we first transform the commonly defined 7-DoF 3D targets to the image domain and decouple them as 2D and 3D attributes to fit the 3D setting. On this basis, the objects are distributed to different feature levels considering their 2D scales and further assigned only according to the 3D centers. In addition, the center-ness is redefined with a 2D Gaussian distribution based on the 3D-center to be compatible with our target formulation.
Experimental results with detailed ablation studies show the efficacy of our approach. For future work, a promising direction is how to better tackle the difficulty of depth and orientation estimation in this ill-posed setting.

{\small
\bibliographystyle{ieee_fullname}
\bibliography{egbib}
}


\clearpage

\twocolumn[{%
\renewcommand\twocolumn[1][]{#1}%
\begin{center}
    \Large
    \textbf{Appendix}
\end{center}
\begin{center}
  \centering
  \includegraphics[width=1.0\linewidth]{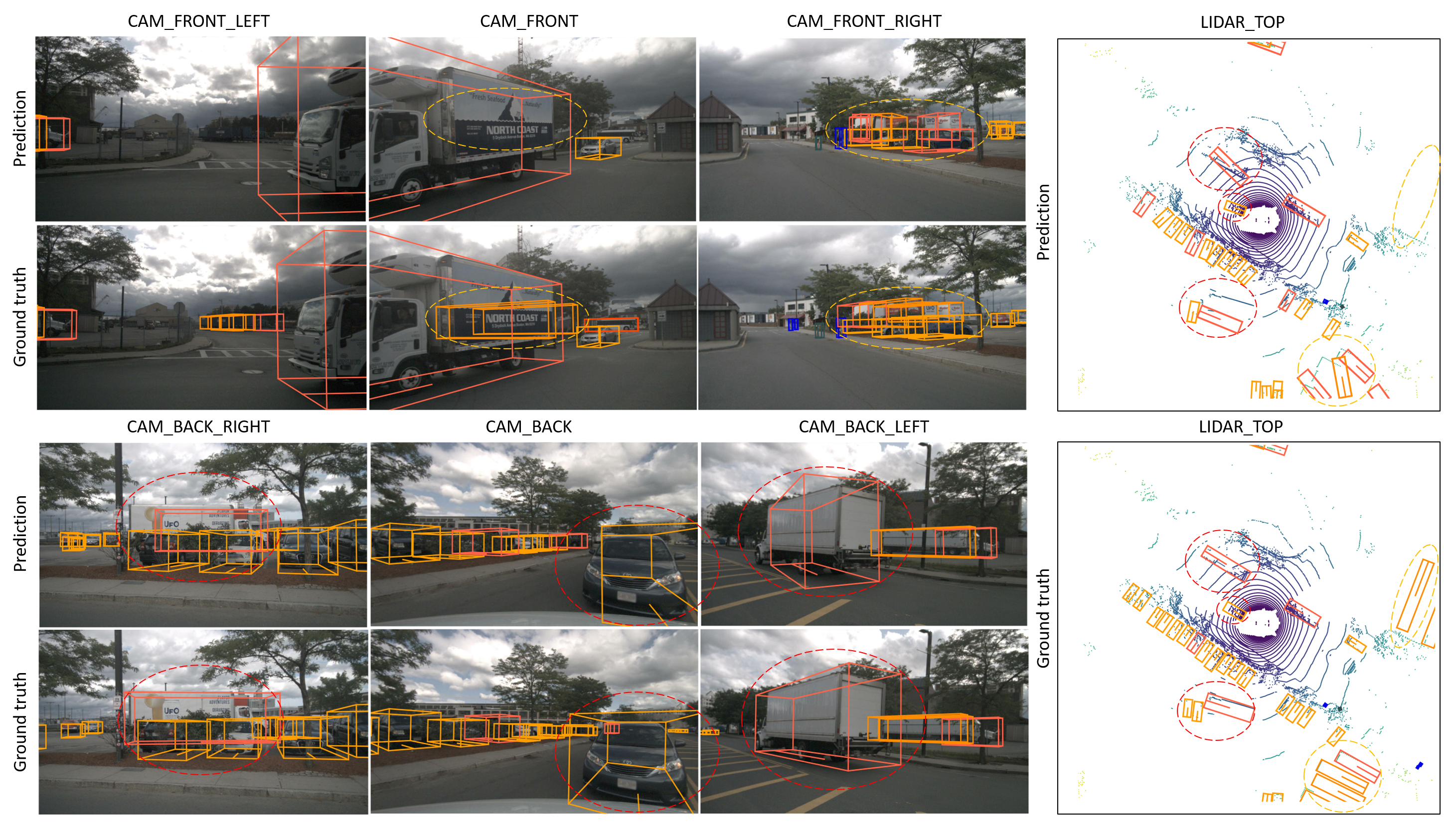}
  \vspace{-4ex}
  \captionof{figure}{Failure cases. As shown in this figure, our detectors perform poorly, especially for occluded and large objects. We use yellow dotted circles to mark the failure case caused by occlusion while use red dotted circles to mark the inaccurate large objects predictions. The former problem is intrinsic, considering the ill-posed property of this task itself. So a direction to improve our method would be how to enhance the detection performance for large objects.}
  \label{fig: failure_case}
  \vspace{1ex}
\end{center}%
}]

\setcounter{section}{0}
\section{Failure Cases}
\vspace{-0.5ex}
In Fig.~\ref{fig: failure_case}, we show some failure cases, mainly focused on the detection of large objects and occluded objects. In the camera view and top view, yellow dotted circles are used to mark the blocked objects that are not successfully detected. Red dotted circles are used to mark the detected large objects with noticeable deviation. The former is mainly manifest in the failure to find the objects behind, while the latter is mainly manifest in the inaccurate estimation of the size and orientation of the objects. The reasons behind the two failure cases are also different. The former is due to the inherent property of the current setting, which is difficult to solve; the latter may be because the receptive field of convolution kernel of the current model is not large enough, resulting in low performance of large object detection. Therefore, the future research direction may be more focused on the solution of the latter.

\newpage
\section{Results on the KITTI Benchmark}
\vspace{-0.5ex}
We provide FCOS3D baseline results on the KITTI benchmark in the follow-up work, PGD~\cite{PGD}. Since the number of samples on KITTI is limited, vanilla FCOS3D cannot achieve outstanding performance. With the basic enhancement of local geometric constraints and customized designs for depth estimation, PGD (can also be termed as FCOS3D++) finally achieves state-of-the-art or competitive performance on various benchmarks under different evaluation metrics. Please refer to the paper~\cite{PGD} for more details.

\end{document}